\documentclass{article}
\usepackage{ijcai25}

\usepackage{times}
\usepackage{soul}
\usepackage{url}
\usepackage[hidelinks]{hyperref}
\usepackage[utf8]{inputenc}
\usepackage[small]{caption}
\usepackage{graphicx}
\usepackage{amsmath}
\usepackage{amsthm}
\usepackage{booktabs}
\usepackage{algorithm}
\usepackage{algorithmic}
\usepackage[switch]{lineno}
\usepackage{multirow}
\usepackage{amssymb}

\title{Query-Based and Unnoticeable Graph Injection Attack from Neighborhood
Perspective}

\author{
    Anonymous Authors
    \affiliations
    Anonymous Affiliation
}

\author{
Chang Liu$^1$
\and
Hai Huang$^2$\and
Yujie Xing$^{2,3}$\And
Xingquan Zuo$^4$\\
\affiliations
$^1$ Beijing University of Posts and Telecommunications\\
\emails
\{changliu, hhuang\}@bupt.edu.cn}

\begin{document}

\maketitle

\begin{abstract}
The robustness of Graph Neural Networks (GNNs) has become an increasingly important topic due to their expanding range of applications. Various attack methods have been proposed to explore the vulnerabilities of GNNs, ranging from Graph Modification Attacks (GMA) to the more practical and flexible Graph Injection Attacks (GIA). However, existing methods face two key challenges: (i) their reliance on surrogate models, which often leads to reduced attack effectiveness due to structural differences and prior biases, and (ii) existing GIA methods often sacrifice attack success rates in undefended settings to bypass certain defense models, thereby limiting their overall effectiveness. To overcome these limitations, we propose QUGIA, a Query-based and Unnoticeable Graph Injection Attack. QUGIA injects nodes by first selecting edges based on victim node connections and then generating node features using a Bayesian framework. This ensures that the injected nodes are similar to the original graph nodes, implicitly preserving homophily and making the attack more unnoticeable. Unlike previous methods, QUGIA does not rely on surrogate models, thereby avoiding performance degradation and achieving better generalization. Extensive experiments on six real-world datasets with diverse characteristics demonstrate that QUGIA achieves unnoticeable attacks and outperforms state-of-the-art attackers. The code will be released upon acceptance.

\end{abstract}

\section{Introduction}

Graph Neural Networks (GNNs), as a representative approach for processing graph-structured data, have shown promising performance in tasks involving relational information \cite{jiang2021could}. GNNs recursively learn feature and topological information through a message-passing paradigm. Nevertheless, this paradigm also leads to their vulnerability \cite{Dai2018Adversarial,zugner2018adversarial}, and many attack methods have been proposed to explore their robustness. Previous studies mainly focus on Graph Modification Attacks (GMA) \cite{2018adversarial,zugner2018adversarial}, which reduce the performance of GNNs by modifying node features and edge connections in the original graph. However, GMA is difficult to implement in practice because attackers often lack the privileges to alter existing data in a graph. To make graph attacks more practical, researchers have proposed Graph Injection Attacks (GIA) \cite{ju2023let,chen2022understanding,tao2021single,zou2021tdgia,wang2020scalable,Sun2020Adversarial}, which reduce the performance of GNNs by injecting malicious nodes.

The settings of GIA can be categorized into white-box and black-box approaches. White-box methods require access to the parameters and architecture of the target model, which is impractical in real-world scenarios. Therefore, we focus on GIA in the black-box setting. Although numerous effective GIA studies have emerged \cite{chen2022understanding,zou2021tdgia,wang2020scalable,Sun2020Adversarial}, this type of attack often disrupts the homophily of the original graph \cite{li2023revisiting,chen2022understanding}. Homophily indicates that neighboring nodes tend to have similar node features or labels, which is a key characteristic of homophilous graphs. Thus, GIA can be effectively defended against by homophily defenders using edge pruning \cite{zhang2020gnnguard}. To address this problem, the Harmonious Adversarial Objective (HAO) \cite{chen2022understanding} enforces GIA to preserve the homophily of the original graph. Nonetheless, according to the no-free-lunch principle, we observe that HAO sacrifices attack performance against undefended models to enhance performance against homophily defenders. Moreover, existing GIA methods mainly rely on surrogate models to generate perturbed graphs. However, different architectures and prior biases pose significant challenges to the generalization ability of the attack method when applied to a new target model. Although recent works \cite{ju2023let,wang2021cluster} attempt to mitigate the dependency on surrogate models by generating perturbed graphs without gradients and using query-based attacks, these methods remain susceptible to homophily-based defenses. These critical flaws raise a significant concern: \textit{How can we design a surrogate-free attack method that minimizes damage to attack performance against undefended models while remaining undetectable by homophily defenders?}

To address the aforementioned challenges, we propose a \textbf{Query-based and Unnoticeable Graph Injection Attack (QUGIA)}, which is entirely independent of surrogate models. Following existing GIA methods, our approach consists of two main components: \textit{feature generation} and \textit{edge generation}. The feature generation component generates the features of the injected nodes, while the edge generation component determines the edges that connect the injected nodes to the original graph nodes. For the feature generation component, we adopt a Bayesian framework to generate the features of the injected nodes, allowing the model to learn from historical Bayesian inference information. We initialize the features of the injected nodes based on those of the victim nodes. Subsequently, we optimize the features of the injected nodes within a fixed perturbation range, $K$, which implicitly preserves the homophily of the original graph. For the edge generation component, previous work generally generates edges from the perspective of node degree, as nodes with lower degrees are less robust \cite{li2023revisiting,zou2021tdgia}. Considering the message-passing paradigm and the homogeneity of the graph, unlike conventional methods, we generate edges from the perspective of the neighbors of the victim nodes. QUGIA avoids the notorious bi-level optimization problem, which involves alternating optimization between edge generation and node generation. Moreover, it implicitly preserves the homogeneity of the graph, making the attack less noticeable. QUGIA provides a novel perspective for the design of GIA attacks. Our contributions are summarized as follows:

\begin{itemize}
\item We investigate query-based GIA that are unnoticeable under defenses while effective against undefended models, which were rarely discussed in previous studies.

\item We propose a query-based attack framework that is independent of specific model assumptions. Our approach introduces a Bayesian optimization-based method for updating injected node features and an injection strategy based on the victim node's neighbors.

\item Extensive experiments on six real-world datasets and six GNN models demonstrate that our attack method outperforms state-of-the-art attackers.

\end{itemize}

\section{Preliminary}
\subsection{Graph Neural Networks}

Consider a graph \( G = (V, E) \), where \( V = \{v_1, \ldots, v_n\} \) represents the set of nodes, and \( E = \{e_1, \ldots, e_m\} \) represents the set of edges. Let \( X \) denote the node feature matrix, with \( X \in \mathbb{R}^{n \times d} \), where \( n \) is the number of nodes and \( d \) is the dimensionality of the node features. The feature vector of a node \( u \) is denoted by \( X_u \). The adjacency matrix is represented as \( A \in \{0, 1\}^{n \times n} \), where \( A_{ij} = 1 \) if there exists an edge between nodes \( i \) and \( j \), and \( A_{ij} = 0 \) otherwise. The set of neighbors of a node \( v \) is denoted by \( \mathcal{N}(v) \). In this paper, we focus on the semi-supervised node classification task. The nodes used during training are denoted as \( V_{\text{train}} \). For each node \( u \) in \( V_{\text{train}} \), there is a corresponding label \( y_u \in Y_L \), where \( Y_L \subseteq Y \) and \( Y = \{1, 2, \ldots, C\} \). \( Y_L \) represents the set of labeled nodes. During the testing phase, the trained GNN model predicts the labels of the nodes in \( V_{\text{test}} = V \setminus V_{\text{train}} \) based on the subgraph \( G_{\text{test}} \).

\subsection{Graph Adversarial Attacks}

During the training phase, the parameters of the GNN model are learned using \( G_{\text{train}} \). The objective of the adversarial attack is to construct a perturbed graph \( G' \) that reduces the classification accuracy of the trained GNN model on \( V_{\text{test}} \) during the testing phase. The general formulation of graph adversarial attacks can be expressed as follows:

\begin{equation}
\min \mathcal{L}_{\text{atk}}(f_{\theta}(G')), \; \text{s.t.} \; \|G' - G\| \leq \Delta,
\label{eq:gnn_loss}
\end{equation}

where \( \theta \) represents the trained parameters of the GNN model, \( \mathcal{L}_{\text{atk}} \) denotes the loss function for the attack (e.g., \( -\mathcal{L}_{\text{train}} \)), and \( \Delta \) constrains the perturbation strength allowed for the attacker.

For GIA, the perturbed adjacency matrix and feature matrix can be expressed as follows:

\begin{equation}
A' = 
\begin{bmatrix} 
A & A_I \\ 
A_I^T & O_I 
\end{bmatrix}, \quad 
V' = 
\begin{bmatrix} 
V \\ 
V_I 
\end{bmatrix},
\label{eq:insert}
\end{equation}

where \( A_I \) denotes the adjacency matrix between the original nodes and the injected nodes, \( O_I \) represents the adjacency matrix among the injected nodes, and \( V_I \) corresponds to the injected nodes. The constraints for GIA attacks can be summarized as:

\begin{equation}
\begin{aligned}
&|V' - V|_0 \leq \Delta, \quad 1 \leq d_u \leq b, \\
&\min(X) \leq X_{u,i} \leq \max(X), \quad \forall i \in \{1, 2, \dots, d\}
\end{aligned}
\label{eq:gnn_constrain}
\end{equation}

where \( \Delta \) represents the maximum number of injected nodes, 
\( d_u \) refers to the degree of the injected node \( u \), 
and \( X_{u,i} \) denotes the feature value of the injected node \( u \) in the \( i \)-th dimension. The terms \( \min(X) \) and \( \max(X) \) represent the minimum and maximum values of all node features in the original graph after normalization. It is worth noting that discrete features are not normalized in this process.

Following previous works \cite{chen2022understanding,zheng2021graph}, this paper primarily focuses on an evasion attack scenario where the attacker is not allowed to modify the model's parameters. The attack operates in an inductive learning setting, meaning that test nodes remain entirely unseen during training. Furthermore, the attack is characterized as non-targeted and black-box. Specifically, the attacker aims to reduce the model's accuracy by causing as many misclassifications as possible across the entire test set within a given attack budget, without targeting specific nodes or classes, and without any knowledge of the model's architecture or parameters.

\section{Proposed Method}

\begin{figure*}[htb]
\centering
\includegraphics[width=1\linewidth]{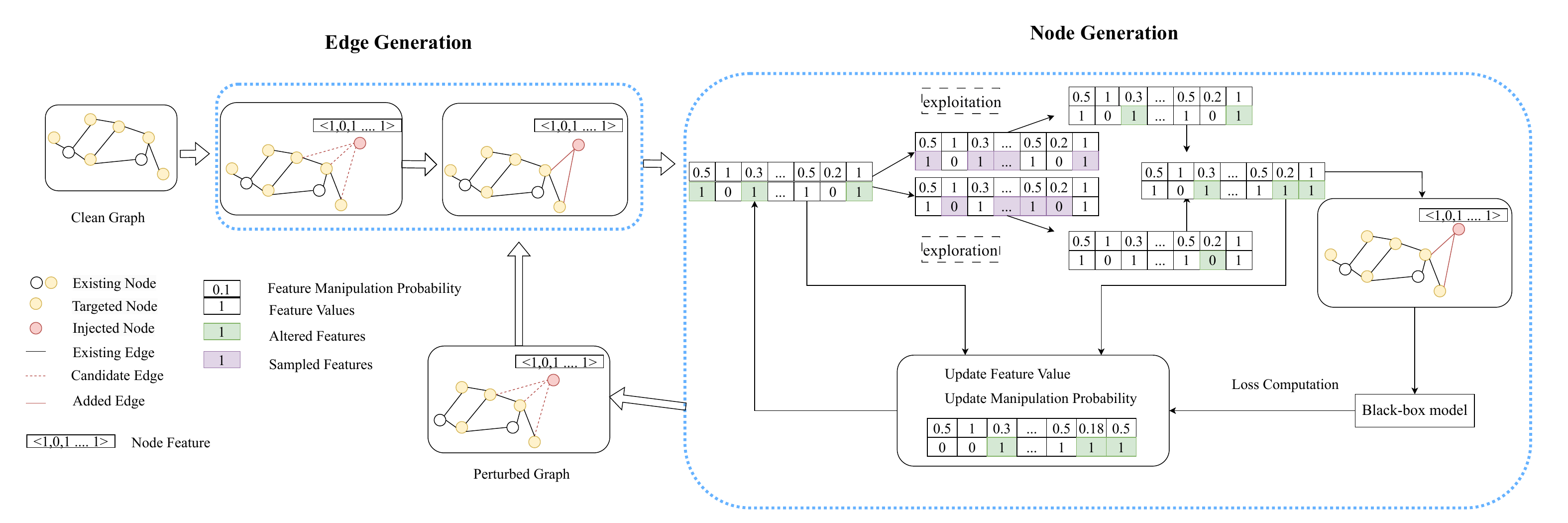}
\caption{QUGIA consists of two components: edge generation and node generation. Edge generation first connects injected nodes to the original graph nodes by considering the neighborhood of the victim nodes, followed by node generation, which optimizes the features of the injected nodes using a novel Bayesian framework.}
\label{fig:framework}
\end{figure*}

In this chapter, we provide a detailed introduction to our proposed attack method, QUGIA. Figure \ref{fig:framework} illustrates the overall framework of QUGIA, which consists of two core components: node generation and edge generation. Node generation optimizes the features of injected nodes using Bayesian optimization, while edge generation determines the connections of injected nodes based on the neighbors of the victim node. In each injection step, QUGIA first performs edge generation to establish the connections for the injected nodes, followed by the optimization of their features. As feature optimization is the primary aspect of the injection process, we will first discuss node generation, followed by edge generation.

\subsection{Edge Generation}

Generating malicious edges is a discrete problem with an extremely large search space due to the numerous possible connections in the graph. Alternating between feature optimization and edge selection leads to the notorious bi-level optimization problem, which is challenging to address within a limited number of queries. Existing works mainly tend to attack low-degree victim nodes because such nodes exhibit lower robustness compared to high-degree nodes \cite{li2023revisiting,zou2021tdgia}. Although attacks from the perspective of node degree have been extensively studied, this approach suggests that the topological structure of the graph can be leveraged to simplify the design of attack methods. Unlike existing approaches, our work designs edge generation by considering the neighbors of the victim node, offering a novel perspective for GIA.

Due to the message-passing paradigm in graphs, changes in a node can affect its neighboring nodes. When a new node $V_u'$ is injected to attack victim node $V_u$, the feature representation of $V_u$ is disrupted, which in turn impacts its neighboring nodes. Therefore, we designed malicious edge generation from the perspective of neighbors, expanding the attack area to $V_u \cup \mathcal{N}(u)$. We only focus on first-order neighbors because the influence between graph nodes diminishes as the distance between them increases. Let $p_u$ to denote the first-order neighbors of node in $V_{test}$, where $p_u = \mathcal{N}_1(u) \cap V_{test}$. The homophily of the graph implies that neighboring nodes (particularly first-order neighbors) are likely to exhibit similar behaviors. If we can implicitly ensure the similarity between $V_u'$ and $V_u$, then the newly injected node and $\mathcal{N}(u)$ are likely to display a higher degree of homophily. This edge generation method not only simplifies the attack method but also implicitly preserves the homophily of the original graph.

We perform attacks in descending order based on $|p_u|$, where $|p_u|$ represents the number of nodes in the set $p_u$. If attacks are performed in ascending order based on $|p_u|$, when $|p_u| = 0$, our attack method degenerates into a single-node attack strategy. When attacking node $u$, $\forall u \in V_{test}$, we connect the malicious new node $X_u'$ to $k$ randomly selected nodes from $p_u$, ensuring that $X_u'$ is always connected to $X_u$.

We utilize the Carlini-Wagner (CW) loss function \cite{xu2019topology} for our loss component, which is expressed as follows:

\begin{equation}
\ell = \max(-\gamma, \:(f_{\theta}(G')_{y_u} - \max_{c \neq y_u} f_{\theta}(G')_c))
\label{eq:cw}
\end{equation}

Here, $\gamma$ represents the confidence score of $\mathcal{L}_{\text{atk}}$. A higher $\gamma$ implies a stronger attack effect but results in a greater distance from the decision boundary. To ensure the attack remains as close to the decision boundary as possible, we set $\gamma = 0.05$. Combining Eq.~(\ref{eq:gnn_loss}) and Eq.~(\ref{eq:cw}), the complete attack loss function is:

\begin{equation}
\ell = \sum\limits_{u \in p_u}(\max(-0.05, (f_{\theta}(x_u')_{y_u} - \max_{c \neq y_u} f_{\theta}(x'_u)_c)))
\label{eq:cw_final}
\end{equation}

To maximize the utilization of the injection node edge budget, we record triplet data \((src, dst, score)\), where \(src \in V_I\), \(dst \in V\), and \(score \in [0, 1]\). There is no existing connection between \(src\) and \(dst\). The \(score\) value, computed using Equation \ref{eq:cw}, represents the distance of the \(dst\) node from the decision boundary, with smaller values indicating a higher likelihood of a successful attack. After completing the attack process, if any injection node edge budget remains, the recorded triplets are sorted in ascending order based on their \(score\) values, and connections are established by prioritizing the pairs with the lowest scores.

\subsection{Node Generation}
Since reliance on surrogate models is eliminated, generating the features of injected nodes cannot depend on neural networks and instead becomes a combinatorial optimization problem. The high dimensionality of graph data features significantly expands the search space, particularly for continuous features, where the search space spans the range $[\min(X), \max(X)]$. To address this issue, we first reduce the dimensionality of the search space and then perform the search within the simplified space. This process presents two primary challenges: (i) determining the magnitude of feature updates and (ii) selecting the dimensions for feature updates.

The magnitude of feature updates refers to the degree of changes applied to the features during each update. In surrogate-based attacks, attackers can utilize gradients to assign varying update sizes at each step. However, in heuristic-based search, gradients are unavailable, and exhaustive search becomes impractical. To address this issue, we define feature updates as the distance between feature values and their boundary values. By doing so, continuous feature values are replaced with discrete, finite values, allowing perturbations within a significantly reduced search space. Specifically, the features of the injected node are defined as $X_u' = s \tilde{X_u} + (1 - s)X_u$, where \( \tilde{X_u} \) is defined in Eq.~(\ref{eq:xu_define}), and \( s \) indicates whether a specific position in the feature vector is flipped, with \( s \in \{0, 1\}^{1 \times d} \).

\begin{equation}
\tilde{X_u} =  \begin{cases}  \min(X) & \text{if } X_u > \frac{\min(X) + \max(X)}{2} \\ \\ \max(X) & \text{otherwise} \end{cases}
\label{eq:xu_define}
\end{equation}

After determining the magnitude of feature updates, we focus on how to select the feature combinations for updates. Specifically, we aim to further reduce the search dimensionality of $s$. Following the inspiration from BBA \cite{pmlr-v162-lee22h}, which first identifies a successful adversarial sample and then searches for unnoticeable perturbations around it, we propose a novel approach from a different perspective. Instead of searching for perturbations around a successful adversarial sample, our goal is to initially identify the most unnoticeable injected node feature $X_{\text{u\_init}}' = X_u$, followed by searching within its neighboring feature space to find $X_u'$ that ensures both a successful attack and unnoticeability. The subsequent challenge is how to define the neighboring feature space. Intuitively, the fewer modifications are made to $X_{\text{u\_init}}'$, the closer it remains to the original feature space, which aligns with the principles of sparse attacks \cite{vo2024brusleattack,croce2019sparse,Narodytska2017Simple}. Based on this intuition, we define a relaxed search budget $K$, where $K \in \mathbb{Z}$, representing positive integers $\{1, 2, 3, \dots\}$. For each injected node feature $X_u'$, we modify only a fixed number of feature dimensions, thereby further reducing the search space. In search problems, the trade-off between exploration and exploitation is inevitable. To manage this trade-off, we employ a power decay strategy, as described in Eq.~(\ref{eq:decad}), where $\lambda_t$ represents the exploration rate at iteration $t$, $B$ denotes the exponential decay speed, and $A$ controls the extent of the initial exploration. Here, $A$ and $B$ are real hyperparameters that can be tuned to balance the trade-off between exploration and exploitation effectively. Larger values of $A$ promote broader initial exploration, while smaller values of $B$ result in faster decay, leading to greater exploitation in later iterations.

\begin{equation}
\lambda_t = A \cdot B^t
\label{eq:decad}
\end{equation}

It is evident that different features contribute variably to the success of the attack, making it reasonable to treat these features differently. To address this, we employ a Bayesian evolutionary algorithm to learn the influence of features from historical data and approximate their impact using a probabilistic framework. Specifically, we utilize a categorical distribution to simulate the influence of different feature dimensions, as selecting various feature combinations corresponds to multiple draws from a set of possible categories. We use the Dirichlet distribution as the prior for the categorical distribution, as it effectively models the probability distribution over each category or combination of categories. The categorical distribution is parameterized by $\theta$, with its prior defined as: $P(\theta; \alpha) := \text{Dir}(\alpha)$, where $\alpha = [\alpha_1, ..., \alpha_d]$ represents the concentration parameters. A uniform distribution is employed with $\alpha_i = 1$. The Bayesian framework is structured into three stages: initialization, sampling, and updating.

\textbf{Initialization of $X_u'$}. We propose a sequential attack method. When attacking a victim node $X_u$, we initialize the features of the injected node $X_u'$ using the features of $X_u$, i.e., $X_u' = X_u$. We randomly select $K$ positions in $s$ and set them to 1. 

\textbf{Sampling of $s_i$}. We perform sampling based on $\theta$ and $s^{(t-1)}$, where $s_i^{(t-1)}$ denotes the value of the $i$-th dimension of the perturbation vector $s$ at the $(t-1)$-th iteration. Eq.~(\ref{eq:sample_1}) focuses on exploring unknown perturbation combinations, while Eq.~(\ref{eq:sample_2}) focuses on exploiting known perturbation combinations.

\begin{equation}
s_1^{(t)}, \ldots, s_{\left\lceil K \cdot \lambda_t \right\rceil}^{(t)} \sim \text{Categorical}(\mathbf{s} \mid \theta^{(t)}, s^{(t-1)} = 0)
\label{eq:sample_1}
\end{equation}

\begin{equation}
s_{\left\lceil K \cdot \lambda_t + 1 \right\rceil}^{(t)}, \ldots, s_{K}^{(t)} \sim \text{Categorical}(\mathbf{s} \mid \theta^{(t)}, s^{(t-1)} = 1)
\label{eq:sample_2}
\end{equation}

\begin{equation}
s^{(t)} = \left[ \bigvee_{i=1}^{K \cdot \lambda_t} s_i^{(t)} \right] \vee \left[ \bigvee_{j=K \cdot \lambda_t + 1}^{K} s_j^{(t)} \right]
\label{eq:sample_or}
\end{equation}

The exact solution for the underlying parameter distribution $\theta$ is typically unattainable. However, since the Dirichlet distribution is a conjugate prior, we can approximate $\theta$ using the expectation of the Dirichlet posterior distribution, which is learned and updated through Bayesian inference over time.

\begin{equation}
    \alpha_{i}^{\text{posterior}} = \alpha_{i}^{\text{prior}} + s_{i}^{(t)}
\label{eq:uppam1}
\end{equation}

\begin{equation}
    \theta^{(t)} = \mathbb{E}_{\theta \sim P(\theta \mid \alpha, u^{(t-1)}, \ell^{(t-1)})}[\theta]
\end{equation}

The posterior concentration parameter \( \alpha_{i}^{\text{posterior}} \) is calculated by adding the observation value \( s_{i}^{(t)} \) from the \( t \)-th iteration to the prior concentration parameter \( \alpha_{i}^{\text{prior}} \). Specifically, \( s_{i}^{(t)} \) is defined as:  $s_{i}^{(t)} = \left(\frac{q_{i}^{(t)} + 0.001}{v_{i}^{(t)} + 0.001}\right)$,  where \( q_{i}^{(t)} \) represents the accumulated importance of the \( i \)-th dimension feature, as described in Eq.~(\ref{eq:q_i}), and \( v_{i}^{(t)} \) represents the accumulated access count for the feature at position \( i \), as described in Eq.~(\ref{eq:v_i}). The loss function at the \( t \)-th iteration of feature optimization is denoted as: $\ell^{(t)} = \mathcal{L}_{\text{atk}}(f_{\theta}(G'))^{(t)}$.

\begin{equation} 
q_{i}^{(t)} = \begin{cases}  
q_{i}^{(t-1)} + 1 & \text{if } \ell^t \geq \ell^{(t-1)} \\
 & \quad \land s_{i}^{(t)} = 1 \land s_{i}^{(t-1)} = 0 \\ 
q_{i}^{(t-1)} & \text{otherwise} 
\end{cases}
\label{eq:q_i}
\end{equation}

\begin{equation}
v_{i}^{(t)} = 
\begin{cases} 
v_{i}^{(t-1)} + 1 & \text{if } s_{i}^{(t)} = 1 \lor s_{i}^{(t-1)} = 1 \\
v_{i}^{(t-1)} & \text{otherwise}
\label{eq:v_i}
\end{cases}
\end{equation}

\textbf{Updating of $s^{(t)}$}. We generate a sample $s^{(t)}$ during the $t$-th feature update, where each $s^{(t)}$ represents a candidate solution. If the new perturbed feature outperforms the perturbed feature from the $(t-1)$-th iteration, we retain the current sample. Otherwise, we retain the sample $s^{(t-1)}$ from the previous iteration. The corresponding formula is as follows:

\begin{equation}
s^{(t)} =
\begin{cases} 
s^{(t)} & \text{if } \ell^{(t)} < \ell^{(t-1)} \\
s^{(t-1)} & \text{otherwise}
\end{cases}
\end{equation}

\subsection{Comprehensive Execution Process of QUGIA}
In this subsection, we provide a detailed explanation of the execution process of QUGIA. The complete execution procedure is presented in Algorithm~\ref{alg:algorithm}. In each injection step, QUGIA first determines the connectivity of the injected node through edge generation (lines 2–3) and then optimizes its features (lines 4–17). After injecting a node, it updates the structural information (line 18) and repeats the process until the node injection budget is exhausted. If any edge injection budget remains, it is allocated through edge generation (line 20).
\begin{algorithm}[htb]
    \caption{QUGIA}
    \label{alg:algorithm}
     \textbf{Input}: Graph $G=(A,X)$, target node set $V_t$, injection node attack budget $\Delta_n$, injection node edge budget $\Delta_e$, maximum iterations $T$, Dirichlet parameters, structural selection score $|p|$, model-predicted label function $f()$, node labels $Y$.
 \\
    \noindent\textbf{Output}: Final adversarial graph $G'$

    \begin{algorithmic}[1]
        \FOR{each $i \in [0, \Delta_n)$}
            \STATE Select victim node $u$ based on sorted scores $|p|$
            \STATE Initialize $A_u'$ by connecting injected nodes to node $u$ and its neighbors

            \WHILE{$t < T$ and $\exists j \in (p_u \cup \{u\}), f(j) \neq Y_j$}
                \STATE Initialize feature matrix $X_u'^{(0)}$
                \STATE Sample perturbation vector $s_i^{(t)}$ according to Equations (\ref{eq:sample_1}--\ref{eq:sample_or})
                \STATE $X_u'^{(t)} \gets s^{(t)} \tilde{X}_u + (1-s^{(t)})X_u$
                \STATE Calculate loss $\ell^t$ for $X_u'^{(t)}$ using Equation~(\ref{eq:cw_final})

                \IF{$\ell^t < \ell^{t-1}$}
                    \vspace{3pt}
                    \STATE $X_u'^{(t)} \gets X_u'^{(t)}$
                    \STATE Update $A'$, $X'$ with $A_u'$, $X_u'^{(t)}$ using ~(\ref{eq:insert})
                \ELSE
                    \STATE $X_u'^{(t)} \gets X_u'^{(t-1)}$
                \ENDIF
                \STATE Update $\lambda_t$ using Equation~(\ref{eq:decad})
                \STATE Update Dirichlet parameters according to Equations (\ref{eq:uppam1}--\ref{eq:v_i})
            \ENDWHILE
            \STATE If $f(u) \neq Y_u$, then for all $m \in \mathcal{N}_1(u) \cap V_{\text{t}}$, the score is updated as $|p_m| = |p_m| - 1$.

        \ENDFOR
        \STATE Allocate remaining edge budget $\Delta_e$
        \STATE \textbf{return} $G' = (A', X')$
    \end{algorithmic}
\end{algorithm}

\section{Experiments}

\begin{table*}[ht]
\centering
\setlength{\tabcolsep}{1mm}
\small
{
\begin{tabular}{l|c|@{\hskip 2mm}cccccccc}
    \hline
    \noalign{\vskip 2pt}
    Datasets & $a$ & TDGIA & TDGIA+HAO & ATDGIA & ATDGIA + HAO & AGIA & AIGA+HAO & G2A2C & QUAGIA\\
    \hline
    \noalign{\vskip 2pt}
    \multirow{3}{*}{Cora} & 0.01 & 0.988$_{\pm 0.001}$ & 0.996$_{\pm 0.001}$ & \underline{0.973$_{\pm 0.002}$} & 0.985$_{\pm 0.001}$ & 0.974$_{\pm 0.003}$ & 0.985$_{\pm 0.002}$ & 0.988$_{\pm 0.002}$ & \textbf{0.956$_{\pm 0.002}$}\\
    & 0.03 & 0.946$_{\pm 0.004}$ & 0.971$_{\pm 0.003}$ & \underline{0.924$_{\pm 0.003}$} & 0.961$_{\pm 0.003}$ & 0.947$_{\pm 0.003}$ & 0.952$_{\pm 0.007}$ & 0.965$_{\pm 0.005}$ & \textbf{0.881$_{\pm 0.004}$} \\
    & 0.05 & 0.899$_{\pm 0.009}$ & 0.929$_{\pm 0.005}$ & \underline{0.898$_{\pm 0.005}$} & 0.924$_{\pm 0.006}$ & 0.920$_{\pm 0.006}$ & 0.923$_{\pm 0.004}$ & 0.953$_{\pm 0.004}$ & \textbf{0.815$_{\pm 0.005}$} \\
    \hline
    \multirow{3}{*}{Citeseer} & 0.01 & 0.991$_{\pm 0.001}$ & 0.998$_{\pm 0.000}$ & \underline{0.981$_{\pm 0.002}$} & 0.991$_{\pm 0.001}$ & 0.982$_{\pm 0.001}$ & 0.992$_{\pm 0.001}$ & 0.993$_{\pm 0.001}$ & \textbf{0.967$_{\pm 0.002}$} \\
    & 0.03 & 0.973$_{\pm 0.002}$ & 0.993$_{\pm 0.002}$ & \underline{0.946$_{\pm 0.003}$} & 0.973$_{\pm 0.002}$ & 0.958$_{\pm 0.004}$ & 0.970$_{\pm 0.002}$ & 0.976$_{\pm 0.002}$ & \textbf{0.900$_{\pm 0.003}$} \\
    & 0.05 & 0.947$_{\pm 0.004}$ & 0.985$_{\pm 0.002}$ & \underline{0.919$_{\pm 0.004}$} & 0.955$_{\pm 0.003}$ & 0.938$_{\pm 0.004}$ & 0.950$_{\pm 0.003}$ & 0.965$_{\pm 0.002}$ & \textbf{0.835$_{\pm 0.004}$} \\
    \hline
    \multirow{3}{*}{GRB-Cora} & 0.01 & 0.975$_{\pm 0.003}$ & 0.993$_{\pm 0.001}$ & \underline{0.967$_{\pm 0.002}$} & 0.976$_{\pm 0.002}$ & 0.969$_{\pm 0.002}$ & 0.975$_{\pm 0.001}$ & 0.991$_{\pm 0.006}$ & \textbf{0.955$_{\pm 0.001}$} \\
    & 0.03 & 0.910$_{\pm 0.003}$ & 0.945$_{\pm 0.003}$ & \underline{0.904$_{\pm 0.003}$} & 0.925$_{\pm 0.004}$ & 0.908$_{\pm 0.004}$ & 0.925$_{\pm 0.004}$ & 0.936$_{\pm 0.010}$ & \textbf{0.848$_{\pm 0.003}$} \\
    & 0.05 & 0.874$_{\pm 0.005}$ & 0.929$_{\pm 0.003}$ & \underline{0.849$_{\pm 0.006}$} & 0.875$_{\pm 0.005}$ & 0.860$_{\pm 0.003}$ & 0.877$_{\pm 0.006}$ & 0.897$_{\pm 0.024}$ & \textbf{0.765$_{\pm 0.006}$} \\
    \hline
    \multirow{3}{*}{GRB-Citeseer} & 0.01 & 0.981$_{\pm 0.004}$ & 0.995$_{\pm 0.001}$ & 0.977$_{\pm 0.002}$ & 0.982$_{\pm 0.001}$ & \underline{0.968$_{\pm 0.004}$} & 0.973$_{\pm 0.002}$ & 0.982$_{\pm 0.002}$ & \textbf{0.957$_{\pm 0.002}$}\\
    & 0.03 & 0.937$_{\pm 0.007}$ & 0.966$_{\pm 0.006}$ & 0.925$_{\pm 0.008}$ & 0.943$_{\pm 0.003}$ & \underline{0.914$_{\pm 0.010}$} & 0.918$_{\pm 0.007}$ & 0.953$_{\pm 0.011}$ & \textbf{0.876$_{\pm 0.004}$}\\
    & 0.05 & 0.891$_{\pm 0.015}$ & 0.930$_{\pm 0.009}$ & 0.883$_{\pm 0.019}$ & 0.900$_{\pm 0.006}$ & \underline{0.862$_{\pm 0.019}$} & 0.865$_{\pm 0.015}$ & 0.927$_{\pm 0.010}$ & \textbf{0.797$_{\pm 0.005}$}\\
    \hline
    \multirow{3}{*}{Pubmed} & 0.01 & 0.990$_{\pm 0.000}$ & 0.994$_{\pm 0.000}$ & 0.984$_{\pm 0.001}$ & 0.988$_{\pm 0.000}$ & \underline{0.982$_{\pm 0.001}$} & 0.990$_{\pm 0.000}$ & 0.986$_{\pm 0.007}$ & \textbf{0.955$_{\pm 0.002}$} \\
    & 0.03 & 0.970$_{\pm 0.000}$ & 0.981$_{\pm 0.001}$ & 0.955$_{\pm 0.001}$ & 0.964$_{\pm 0.001}$ & \underline{0.951$_{\pm 0.003}$} & 0.969$_{\pm 0.001}$ & 0.979$_{\pm 0.006}$ & \textbf{0.879$_{\pm 0.005}$} \\
    & 0.05 & 0.950$_{\pm 0.001}$ & 0.965$_{\pm 0.001}$ & 0.927$_{\pm 0.002}$ & 0.938$_{\pm 0.002}$ & \underline{0.924$_{\pm 0.005}$} & 0.951$_{\pm 0.002}$ & 0.974$_{\pm 0.010}$ & \textbf{0.820$_{\pm 0.007}$} \\
    \hline
    \multirow{3}{*}{Arxiv} & 0.01 & 0.983$_{\pm 0.002}$ & 0.996$_{\pm 0.000}$ & 0.978$_{\pm 0.001}$ & 0.982$_{\pm 0.001}$ & \underline{0.972$_{\pm 0.001}$} & 0.977$_{\pm 0.001}$ & 0.982$_{\pm 0.006}$ & \textbf{0.943$_{\pm 0.001}$} \\
    & 0.03 & 0.943$_{\pm 0.005}$ & 0.971$_{\pm 0.003}$ & 0.937$_{\pm 0.003}$ & 0.943$_{\pm 0.003}$ & \underline{0.926$_{\pm 0.005}$} & 0.934$_{\pm 0.004}$ & 0.933$_{\pm 0.014}$ & \textbf{0.852$_{\pm 0.002}$} \\
    & 0.05 & 0.900$_{\pm 0.007}$ & 0.934$_{\pm 0.003}$ & 0.893$_{\pm 0.007}$ & 0.907$_{\pm 0.003}$ & \underline{0.881$_{\pm 0.006}$} & 0.889$_{\pm 0.005}$ & 0.903$_{\pm 0.018}$ & \textbf{0.778$_{\pm 0.003}$} \\
    \hline
  \end{tabular}
}
\caption{Average performance of various attack methods on undefended models, where lower values indicate better performance.}
\label{tab:no_defense}
\end{table*}

\begin{table*}[!ht]
\centering
\setlength{\tabcolsep}{1mm}
\small
{
\begin{tabular}{l|c|cccccccc}
    \hline
    \noalign{\vskip 2pt}
    Datasets & $a$ & TDGIA & TDGIA+HAO & ATDGIA & ATDGIA + HAO & AGIA & AIGA+HAO & G2A2C & QUAGIA\\
    \hline
    \noalign{\vskip 2pt}
    \multirow{3}{*}{Cora} & 0.01 & 0.998$_{\pm 0.000}$ & 0.992$_{\pm 0.003}$ & 0.996$_{\pm 0.001}$ & 0.982$_{\pm 0.002}$ & 0.993$_{\pm 0.002}$ & \underline{0.980$_{\pm 0.002}$} & 0.992$_{\pm 0.005}$ & \textbf{0.956$_{\pm 0.004}$} \\
    & 0.03 & 0.991$_{\pm 0.003}$ & 0.965$_{\pm 0.002}$ & 0.987$_{\pm 0.003}$ & 0.955$_{\pm 0.005}$ & 0.985$_{\pm 0.004}$ & \underline{0.938$_{\pm 0.007}$} & 0.992$_{\pm 0.009}$ & \textbf{0.880$_{\pm 0.005}$} \\
    & 0.05 & 0.984$_{\pm 0.004}$ & 0.909$_{\pm 0.005}$ & 0.983$_{\pm 0.004}$ & \underline{0.908$_{\pm 0.004}$} & 0.981$_{\pm 0.003}$ & 0.910$_{\pm 0.006}$ & 0.982$_{\pm 0.008}$ & \textbf{0.818$_{\pm 0.004}$} \\
    \hline
    \multirow{3}{*}{Citeseer} & 0.01 & 1.000$_{\pm 0.000}$ & 0.997$_{\pm 0.000}$ & 0.998$_{\pm 0.000}$ & \underline{0.989$_{\pm 0.001}$} & 0.998$_{\pm 0.000}$ & 0.992$_{\pm 0.001}$ & 0.999$_{\pm 0.000}$ & \textbf{0.965$_{\pm 0.001}$} \\
    & 0.03 & 0.999$_{\pm 0.000}$ & 0.989$_{\pm 0.001}$ & 0.994$_{\pm 0.001}$ & 0.971$_{\pm 0.002}$ & 0.995$_{\pm 0.000}$ & \underline{0.966$_{\pm 0.003}$} & 0.999$_{\pm 0.000}$ & \textbf{0.900$_{\pm 0.001}$}\\
    & 0.05 & 0.994$_{\pm 0.003}$ & 0.975$_{\pm 0.002}$ & 0.989$_{\pm 0.003}$ & 0.951$_{\pm 0.003}$ & 0.992$_{\pm 0.002}$ & \underline{0.942$_{\pm 0.003}$} & 0.997$_{\pm 0.002}$ & \textbf{0.839$_{\pm 0.005}$}\\
    \hline
    \multirow{3}{*}{GRB-Cora} & 0.01 & 0.997$_{\pm 0.002}$ & 0.988$_{\pm 0.003}$ &  0.990$_{\pm 0.002}$ & 0.964$_{\pm 0.003}$ & 0.995$_{\pm 0.003}$ & \underline{0.962$_{\pm 0.003}$} & 0.995$_{\pm 0.002}$ & \textbf{0.956$_{\pm 0.003}$}\\
    & 0.03 & 0.992$_{\pm 0.002}$ & 0.920$_{\pm 0.002}$ & 0.989$_{\pm 0.003}$ & \underline{0.889$_{\pm 0.005}$} & 0.993$_{\pm 0.001}$ & 0.890$_{\pm 0.005}$ & 0.924$_{\pm 0.012}$ & \textbf{0.831$_{\pm 0.006}$}\\
    & 0.05 & 0.989$_{\pm 0.003}$ & 0.902$_{\pm 0.004}$ & 0.985$_{\pm 0.002}$ & \underline{0.819$_{\pm 0.007}$} & 0.987$_{\pm 0.003}$ & 0.828$_{\pm 0.006}$ & 0.904$_{\pm 0.015}$ & \textbf{0.742$_{\pm 0.005}$}\\
    \hline
    \multirow{3}{*}{GRB-Citeseer} & 0.01 & 1.000$_{\pm 0.000}$ & 0.995$_{\pm 0.000}$ & 0.996$_{\pm 0.000}$ & 0.978$_{\pm 0.002}$ & 0.998$_{\pm 0.000}$ & \underline{0.969$_{\pm 0.004}$} & 0.982$_{\pm 0.006}$ & \textbf{0.956$_{\pm 0.003}$} \\
    & 0.03 & 0.999$_{\pm 0.000}$ & 0.963$_{\pm 0.005}$ & 0.989$_{\pm 0.001}$ & 0.931$_{\pm 0.008}$ & 0.994$_{\pm 0.001}$ & \underline{0.910$_{\pm 0.013}$} &  0.946$_{\pm 0.008}$ & \textbf{0.876$_{\pm 0.007}$}\\
    & 0.05 & 0.996$_{\pm 0.002}$ & 0.921$_{\pm 0.008}$ & 0.986$_{\pm 0.001}$ & 0.881$_{\pm 0.009}$ & 0.992$_{\pm 0.001}$ & \underline{0.856$_{\pm 0.017}$} & 0.908$_{\pm 0.015}$ & \textbf{0.799$_{\pm 0.004}$}\\
    \hline
    \multirow{3}{*}{Pubmed} & 0.01 & 0.999$_{\pm 0.000}$ & 0.988$_{\pm 0.003}$ & 0.997$_{\pm 0.002}$ & 0.985$_{\pm 0.002}$ & 0.998$_{\pm 0.002}$ & \underline{0.980$_{\pm 0.003}$} & 0.986$_{\pm 0.009}$ & \textbf{0.961$_{\pm 0.001}$} \\
    & 0.03 & 0.995$_{\pm 0.003}$ & 0.971$_{\pm 0.003}$ & 0.994$_{\pm 0.002}$ & \underline{0.953$_{\pm 0.002}$} & 0.996$_{\pm 0.002}$ & \underline{0.953$_{\pm 0.004}$} & 0.964$_{\pm 0.007}$ & \textbf{0.899$_{\pm 0.002}$}\\
    & 0.05 & 0.996$_{\pm 0.006}$ & 0.951$_{\pm 0.003}$ & 0.989$_{\pm 0.004}$ & \underline{0.925$_{\pm 0.004}$} & 0.994$_{\pm 0.003}$ & 0.930$_{\pm 0.003}$ & 0.935$_{\pm 0.017}$ & \textbf{0.833$_{\pm 0.004}$}\\
    \hline
    \multirow{3}{*}{Arxiv} & 0.01 & 0.999$_{\pm 0.000}$ & 0.995$_{\pm 0.000}$ & 0.995$_{\pm 0.001}$ & 0.975$_{\pm 0.001}$ & 0.998$_{\pm 0.000}$ & \underline{0.970$_{\pm 0.002}$} & 0.974$_{\pm 0.012}$ & \textbf{0.941$_{\pm 0.002}$}\\
    & 0.03 & 0.996$_{\pm 0.000}$ & 0.961$_{\pm 0.004}$ & 0.984$_{\pm 0.002}$ & 0.918$_{\pm 0.004}$ & 0.995$_{\pm 0.000}$ & \underline{0.908$_{\pm 0.007}$} & 0.927$_{\pm 0.014}$ & \textbf{0.845$_{\pm 0.001}$}\\
    & 0.05 & 0.989$_{\pm 0.002}$ & 0.909$_{\pm 0.004}$ & 0.976$_{\pm 0.003}$ & 0.872$_{\pm 0.007}$ & 0.990$_{\pm 0.001}$ & \underline{0.856$_{\pm 0.007}$} & 0.873$_{\pm 0.021}$ & \textbf{0.765$_{\pm 0.003}$}\\
    \hline
  \end{tabular}
}
\caption{Average performance of various attack methods on defended models, where lower values indicate better performance.}
\label{tab:defense}
\end{table*}

\subsection{Datasets}

We evaluate our approach on six datasets, covering both discrete and continuous features. The discrete datasets include Cora and Citeseer \cite{yang2016revisiting}, while the continuous datasets comprise Pubmed \cite{sen2008collective}, the GRB-redefined versions of Cora and Citeseer \cite{zheng2021graph}, and the arXiv dataset from OGB \cite{hu2020open}. We adopt a data-splitting strategy similar to those used in prior studies \cite{chen2022understanding,zheng2021graph}. Detailed descriptions of the datasets are provided in the \textit{Supplementary Material}.

\subsection{Baseline}

The study of node insertion attacks in black-box settings remains relatively unexplored. To provide a comprehensive evaluation, we selected several representative methods as baselines. These include TDGIA \cite{zou2021tdgia}, the state-of-the-art method for node insertion attacks; ATDGIA \cite{chen2022understanding}, a variant of TDGIA; and AGIA \cite{chen2022understanding}, which adopts a Gradient-Driven Injection strategy. Additionally, we incorporated HAO \cite{chen2022understanding} into TDGIA, ATDGIA, and AGIA to thoroughly assess the unnoticeability of the attacks. Since these methods require gradients from the victim model, we set the surrogate model to have the same architecture as the victim model, with a fixed random seed of 666. Furthermore, we included G2A2C \cite{ju2023let}, a recent query-based attack that leverages reinforcement learning. To ensure a fair comparison, G2A2C was configured to operate under the same maximum number of single-node queries as our proposed method.

\subsection{Evaluation Protocol}

To ensure a comprehensive comparison, we selected three widely used GNN models without defense mechanisms and three GNN models with defense mechanisms. For models without defenses, we included GCN \cite{GCN17}, GAT \cite{GAT18}, and APPNP \cite{appnp2019}. For models with defenses, following prior work \cite{chen2022understanding}, we selected Guard \cite{zhang2020gnnguard}, EGuard, and RGAT \cite{chen2022understanding}.

The number of inserted nodes is defined as \( |V_I| = a|V| \), where $a$ represents the percentage of the total number of nodes in the clean graph. To evaluate the performance of various attack methods under different attack constraints, we set $a$ to 1\%, 3\%, and 5\%. The maximum degree of the inserted nodes is constrained by the average degree of the corresponding graph dataset.

All attack methods were executed using five different random seeds, and we report both the mean and variance of the results across these five runs. The average performance of the attack methods is summarized in Table~\ref{tab:no_defense} and Table~\ref{tab:defense}. In the no-defense setting, we present the average results across GCN, GAT, and APPNP, while in the defense setting, the results are averaged over Guard, EGuard, and RGAT. For additional details, including experimental configurations, comprehensive results, and sensitivity analyses, please refer to the \textit{Supplementary Material}.

\subsection{Performance Comparison}
\label{Comparison}
In Table~\ref{tab:no_defense} and Table~\ref{tab:defense}, we present the test set classification accuracy of GNN models under various attack methods, with lower values indicating stronger attack performance. The best results are highlighted in bold, while the second-best results are underlined. These experimental results demonstrate that our attack method achieves superior performance on both defended and undefended GNN models. Specifically, our method outperforms AGIA by nearly 5\% on average and up to 10\% in the best case on undefended models. On defended models, our method performs significantly better on discrete datasets compared to other methods. This indicates that gradient-based approaches struggle with discrete data, as gradients are more suitable for optimizing features in continuous data. Although G2A2C reduces reliance on surrogate models, its focus on single-node attacks hinders the effective allocation of the attack budget across multiple nodes, leading to weaker performance.

Additionally, we found that the performance of gradient-based attack methods drops significantly on defended models when the HAO module is removed. Our method achieves up to a 10\% improvement compared to the effective AGIA+HAO attack method on defended models. Interestingly, attack methods with the HAO module generally perform worse on undefended models but better on defended ones. This suggests that the HAO module sacrifices some maliciousness to preserve homophily, making attacks less noticeable. Compared to HAO, our attack method implicitly preserves the homophily of the graph, making it harder to detect using homophily-based defense mechanisms. Furthermore, our attack method maintains excellent performance even on undefended GNN models.

Moreover, we observed that APPNP exhibits surprising robustness to GIA attacks, consistent with findings in previous work \cite{zhang2023minimum}. This robustness is likely due to the residual connections in APPNP, which preserve the original feature information. Our experiments show that different models exhibit varying levels of robustness to GIA attacks. Exploring effective GIA attack and defense mechanisms remains an important research direction.

\subsection{Homophily Analysis}

\begin{figure}[htbp]
\centering
\includegraphics[width=1.\linewidth]{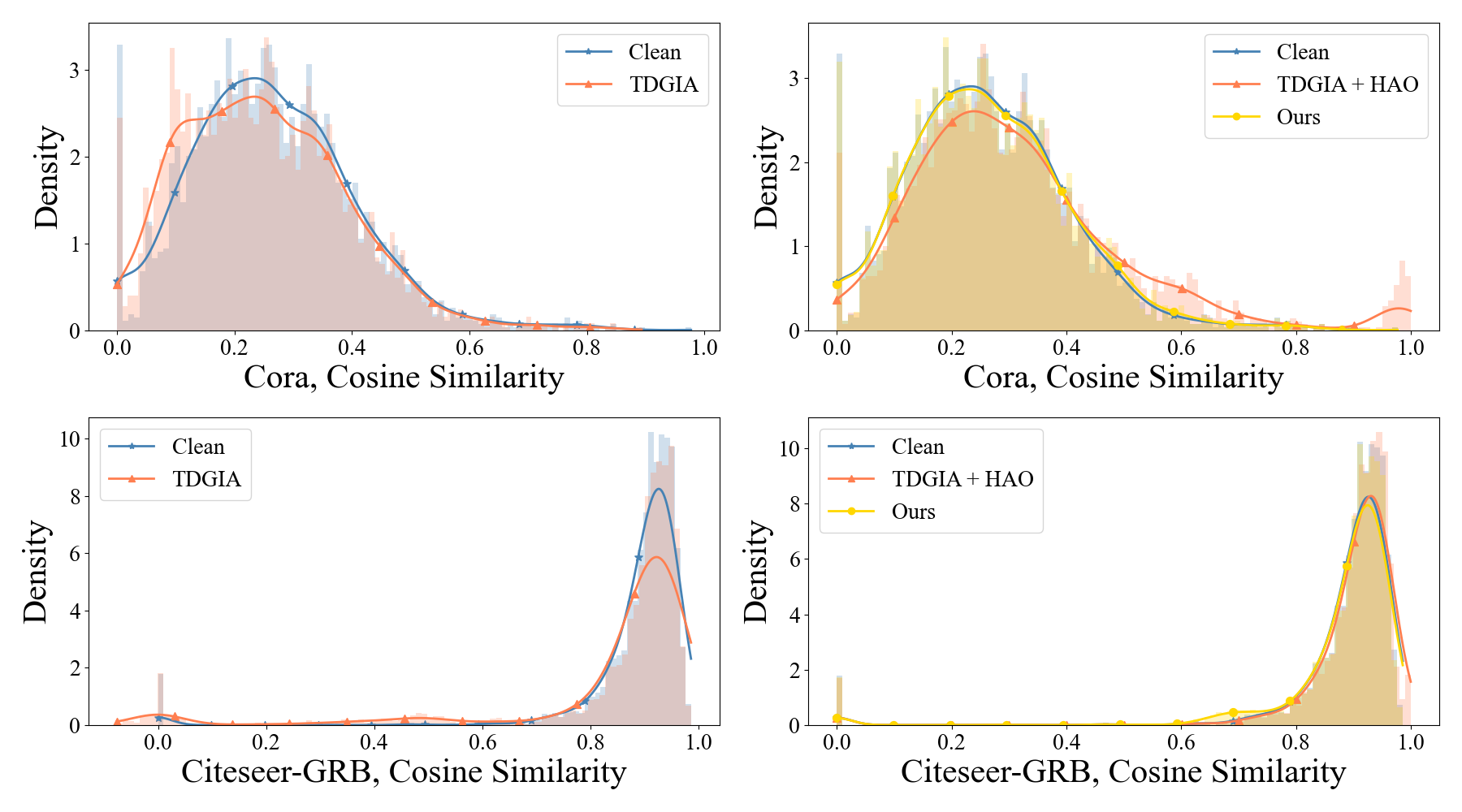}
\caption{Node similarity distributions under different attacks for the Cora and GRB-Citeseer datasets. The x-axis represents cosine similarity, while the y-axis depicts the density of each similarity value.}
\label{fig:node_sim}
\end{figure}

To further investigate the differences between our method and HAO, we compared our approach with the best baseline under defense models, TDGIA+HAO, on the Cora and GRB-Citeseer datasets with $\alpha = 0.03$ and a fixed seed of 0. Following the discussion in previous work \cite{chen2022understanding}, we utilize node similarity to analyze the homophily distribution. Node similarity is defined as the similarity between the features of node $u$ and the aggregated features of its neighbors. As illustrated in Figure \ref{fig:node_sim}, when TDGIA is applied without HAO, the node similarity of the perturbed graph deviates significantly from the distribution of the original clean graph. However, the addition of HAO effectively reduces this deviation. Furthermore, we observed that HAO is less effective on datasets with discrete features compared to those with continuous features, likely due to its reliance on surrogate model gradients.

Similar to HAO, our method also brings the homophily distribution of the perturbed graph closer to that of the original graph. Our attack method preserves the homophily of both discrete and continuous features. In GRB-Citeseer, we shifted a small number of nodes with a similarity of around 0.9 to approximately 0.7. Experimental results suggest that this shift is less likely to be detected by homophily defenders. Balancing the protection of homogeneity while enhancing the performance of attack methods against undefended models remains a challenge.

\subsection{Ablation Analysis}

Refer to the \textit{Supplementary Material} for details on the ablation analysis. Detailed experimental results for Table~\ref{tab:no_defense} and Table~\ref{tab:defense} are also provided in the \textit{Supplementary Material}.

\section{Related Work}

With the widespread adoption of GNNs, their robustness against adversarial attacks has gained increasing attention. While early studies mainly focused on GMA, constraints such as user permissions have led to a shift toward GIA, which has emerged as a more practical and effective attack approach. NIPA \cite{Sun2020Adversarial} achieves its malicious intent by generating a batch of random nodes and injecting them into the existing graph. AFGSM \cite{wang2020scalable} employs the fast gradient sign method to inject malicious nodes, enabling attacks on large-scale graphs. G-NIA \cite{tao2021single} utilizes neural networks to learn and generate new nodes and edges, which are then injected into the original graph to launch attacks. TDGIA \cite{zou2021tdgia} proposes a method for detecting vulnerable nodes in the graph topology to identify attack targets and introduces a smooth feature generation approach to facilitate attacks. AGIA \cite{chen2022understanding} utilizes surrogate model gradients to optimize edge weights and inject node features. HAO \cite{chen2022understanding} offers a plug-in method to enhance the unnoticeability of attacks, utilizing the homophily ratio of nodes as an unnoticeability constraint to further improve the stealthiness of adversarial attacks. However, the aforementioned methods all rely on surrogate models, and discrepancies between the target model and the surrogate model may lead to degraded attack performance. G2A2C \cite{ju2023let} formulates the attack process as a Markov Decision Process (MDP) and utilizes reinforcement learning to optimize an injection model based on query-based learning. While G2A2C alleviates dependency on surrogate model gradients, it does not consider the unnoticeability of the attack and still depends on surrogate models.

\section{Conclusion}

In this work, we propose a neighbor perspective-based attack method that uses a Bayesian framework to generate features for injected nodes, completely eliminating the dependence on surrogate models. Our attack method implicitly preserves the homophily of the original graph, ensuring excellent attack performance under both homophily defenders and undefended models. Extensive Experimental results show that our method outperforms existing state-of-the-art attack methods.

\bibliographystyle{named}
\bibliography{ijcai25}

\begin{thebibliography}{}

\bibitem[\protect\citeauthoryear{Chen \bgroup \em et al.\egroup
  }{2022}]{chen2022understanding}
Yongqiang Chen, Han Yang, Yonggang Zhang, MA~KAILI, Tongliang Liu, Bo~Han, and
  James Cheng.
\newblock Understanding and improving graph injection attack by promoting
  unnoticeability.
\newblock In {\em International Conference on Learning Representations}, 2022.

\bibitem[\protect\citeauthoryear{Croce and Hein}{2019}]{croce2019sparse}
Francesco Croce and Matthias Hein.
\newblock Sparse and imperceivable adversarial attacks.
\newblock In {\em Proceedings of the IEEE/CVF international conference on
  computer vision}, pages 4724--4732, 2019.

\bibitem[\protect\citeauthoryear{Dai \bgroup \em et al.\egroup
  }{2018}]{Dai2018Adversarial}
Hanjun Dai, Hui Li, Tian Tian, Xin Huang, Lin Wang, Jun Zhu, and Le~Song.
\newblock Adversarial attack on graph structured data.
\newblock In {\em International conference on machine learning}, pages
  1115--1124. PMLR, 2018.

\bibitem[\protect\citeauthoryear{Gasteiger \bgroup \em et al.\egroup
  }{2019}]{appnp2019}
Johannes Gasteiger, Aleksandar Bojchevski, and Stephan Günnemann.
\newblock Combining neural networks with personalized pagerank for
  classification on graphs.
\newblock In {\em International Conference on Learning Representations}, 2019.

\bibitem[\protect\citeauthoryear{Hu \bgroup \em et al.\egroup
  }{2020}]{hu2020open}
Weihua Hu, Matthias Fey, Marinka Zitnik, Yuxiao Dong, Hongyu Ren, Bowen Liu,
  Michele Catasta, and Jure Leskovec.
\newblock Open graph benchmark: Datasets for machine learning on graphs.
\newblock {\em Advances in neural information processing systems},
  33:22118--22133, 2020.

\bibitem[\protect\citeauthoryear{Jiang \bgroup \em et al.\egroup
  }{2021}]{jiang2021could}
Dejun Jiang, Zhenxing Wu, Chang-Yu Hsieh, Guangyong Chen, Ben Liao, Zhe Wang,
  Chao Shen, Dongsheng Cao, Jian Wu, and Tingjun Hou.
\newblock Could graph neural networks learn better molecular representation for
  drug discovery? a comparison study of descriptor-based and graph-based
  models.
\newblock {\em Journal of cheminformatics}, 13:1--23, 2021.

\bibitem[\protect\citeauthoryear{Ju \bgroup \em et al.\egroup
  }{2023}]{ju2023let}
Mingxuan Ju, Yujie Fan, Chuxu Zhang, and Yanfang Ye.
\newblock Let graph be the go board: gradient-free node injection attack for
  graph neural networks via reinforcement learning.
\newblock In {\em Proceedings of the AAAI Conference on Artificial
  Intelligence}, volume~37, pages 4383--4390, 2023.

\bibitem[\protect\citeauthoryear{Kipf and Welling}{2017}]{GCN17}
Thomas~N. Kipf and Max Welling.
\newblock Semi-supervised classification with graph convolutional networks.
\newblock In {\em International Conference on Learning Representations}, 2017.

\bibitem[\protect\citeauthoryear{Lee \bgroup \em et al.\egroup
  }{2022}]{pmlr-v162-lee22h}
Deokjae Lee, Seungyong Moon, Junhyeok Lee, and Hyun~Oh Song.
\newblock Query-efficient and scalable black-box adversarial attacks on
  discrete sequential data via bayesian optimization.
\newblock In {\em International Conference on Machine Learning}, pages
  12478--12497. PMLR, 2022.

\bibitem[\protect\citeauthoryear{Li \bgroup \em et al.\egroup
  }{2023}]{li2023revisiting}
Kuan Li, Yang Liu, Xiang Ao, and Qing He.
\newblock Revisiting graph adversarial attack and defense from a data
  distribution perspective.
\newblock In {\em The Eleventh International Conference on Learning
  Representations}, 2023.

\bibitem[\protect\citeauthoryear{Narodytska and
  Kasiviswanathan}{2017}]{Narodytska2017Simple}
Nina Narodytska and Shiva~Prasad Kasiviswanathan.
\newblock Simple black-box adversarial attacks on deep neural networks.
\newblock In {\em CVPR Workshops}, volume~2, 2017.

\bibitem[\protect\citeauthoryear{Sen \bgroup \em et al.\egroup
  }{2008}]{sen2008collective}
Prithviraj Sen, Galileo Namata, Mustafa Bilgic, Lise Getoor, Brian Galligher,
  and Tina Eliassi-Rad.
\newblock Collective classification in network data.
\newblock {\em AI magazine}, 29(3):93--93, 2008.

\bibitem[\protect\citeauthoryear{Sun \bgroup \em et al.\egroup
  }{2020}]{Sun2020Adversarial}
Yiwei Sun, Suhang Wang, Xianfeng Tang, Tsung-Yu Hsieh, and Vasant Honavar.
\newblock Adversarial attacks on graph neural networks via node injections: A
  hierarchical reinforcement learning approach.
\newblock In {\em Proceedings of the Web Conference 2020}, pages 673--683,
  2020.

\bibitem[\protect\citeauthoryear{Tao \bgroup \em et al.\egroup
  }{2021}]{tao2021single}
Shuchang Tao, Qi~Cao, Huawei Shen, Junjie Huang, Yunfan Wu, and Xueqi Cheng.
\newblock Single node injection attack against graph neural networks.
\newblock In {\em Proceedings of the 30th ACM International Conference on
  Information \& Knowledge Management}, pages 1794--1803, 2021.

\bibitem[\protect\citeauthoryear{Veličković \bgroup \em et al.\egroup
  }{2018}]{GAT18}
Petar Veličković, Guillem Cucurull, Arantxa Casanova, Adriana Romero, Pietro
  Liò, and Yoshua Bengio.
\newblock Graph attention networks.
\newblock In {\em International Conference on Learning Representations}, 2018.

\bibitem[\protect\citeauthoryear{Vo \bgroup \em et al.\egroup
  }{2024}]{vo2024brusleattack}
Quoc~Viet Vo, Ehsan Abbasnejad, and Damith Ranasinghe.
\newblock {BRUSLEATTACK}: A {QUERY}-{EFFICIENT} {SCORE}- {BASED} {BLACK}-{BOX}
  {SPARSE} {ADVERSARIAL} {ATTACK}.
\newblock In {\em The Twelfth International Conference on Learning
  Representations}, 2024.

\bibitem[\protect\citeauthoryear{Wang \bgroup \em et al.\egroup
  }{2020}]{wang2020scalable}
Jihong Wang, Minnan Luo, Fnu Suya, Jundong Li, Zijiang Yang, and Qinghua Zheng.
\newblock Scalable attack on graph data by injecting vicious nodes.
\newblock {\em Data Mining and Knowledge Discovery}, 34:1363--1389, 2020.

\bibitem[\protect\citeauthoryear{Wang \bgroup \em et al.\egroup
  }{2021}]{wang2021cluster}
Zhengyi Wang, Zhongkai Hao, Ziqiao Wang, Hang Su, and Jun Zhu.
\newblock Cluster attack: Query-based adversarial attacks on graphs with
  graph-dependent priors.
\newblock {\em arXiv preprint arXiv:2109.13069}, 2021.

\bibitem[\protect\citeauthoryear{Xu \bgroup \em et al.\egroup
  }{2019}]{xu2019topology}
Kaidi Xu, Hongge Chen, Sijia Liu, Pin-Yu Chen, Tsui-Wei Weng, Mingyi Hong, and
  Xue Lin.
\newblock Topology attack and defense for graph neural networks: An
  optimization perspective.
\newblock {\em arXiv}, 2019.

\bibitem[\protect\citeauthoryear{Yang \bgroup \em et al.\egroup
  }{2016}]{yang2016revisiting}
Zhilin Yang, William Cohen, and Ruslan Salakhudinov.
\newblock Revisiting semi-supervised learning with graph embeddings.
\newblock In {\em International conference on machine learning}, pages 40--48.
  PMLR, 2016.

\bibitem[\protect\citeauthoryear{Zhang and Zitnik}{2020}]{zhang2020gnnguard}
Xiang Zhang and Marinka Zitnik.
\newblock Gnnguard: Defending graph neural networks against adversarial
  attacks.
\newblock {\em Advances in neural information processing systems},
  33:9263--9275, 2020.

\bibitem[\protect\citeauthoryear{Zhang \bgroup \em et al.\egroup
  }{2023}]{zhang2023minimum}
Mengmei Zhang, Xiao Wang, Chuan Shi, Lingjuan Lyu, Tianchi Yang, and Junping
  Du.
\newblock Minimum topology attacks for graph neural networks.
\newblock In {\em Proceedings of the ACM Web Conference 2023}, pages 630--640,
  2023.

\bibitem[\protect\citeauthoryear{Zheng \bgroup \em et al.\egroup
  }{2021}]{zheng2021graph}
Qinkai Zheng, Xu~Zou, Yuxiao Dong, Yukuo Cen, Da~Yin, Jiarong Xu, Yang Yang,
  and Jie Tang.
\newblock Graph robustness benchmark: Benchmarking the adversarial robustness
  of graph machine learning.
\newblock In {\em Thirty-fifth Conference on Neural Information Processing
  Systems Datasets and Benchmarks Track (Round 2)}, 2021.

\bibitem[\protect\citeauthoryear{Zou \bgroup \em et al.\egroup
  }{2021}]{zou2021tdgia}
Xu~Zou, Qinkai Zheng, Yuxiao Dong, Xinyu Guan, Evgeny Kharlamov, Jialiang Lu,
  and Jie Tang.
\newblock Tdgia: Effective injection attacks on graph neural networks.
\newblock In {\em Proceedings of the 27th ACM SIGKDD Conference on Knowledge
  Discovery \& Data Mining}, pages 2461--2471, 2021.

\bibitem[\protect\citeauthoryear{Z{\"u}gner \bgroup \em et al.\egroup
  }{2018}]{zugner2018adversarial}
Daniel Z{\"u}gner, Amir Akbarnejad, and Stephan G{\"u}nnemann.
\newblock Adversarial attacks on neural networks for graph data.
\newblock In {\em Proceedings of the 24th ACM SIGKDD international conference
  on knowledge discovery \& data mining}, pages 2847--2856, 2018.

\bibitem[\protect\citeauthoryear{Zügner and
  Günnemann}{2019}]{2018adversarial}
Daniel Zügner and Stephan Günnemann.
\newblock Adversarial attacks on graph neural networks via meta learning.
\newblock In {\em International Conference on Learning Representations}, 2019.

\end{thebibliography}

\end{document}